\documentclass{bmvc2k}

\usepackage{booktabs}
\usepackage{algpseudocode}
\usepackage{algorithm}
\usepackage{mathtools}
\usepackage{tikz}
\usepackage{amssymb}

\newcommand{\blue}[1]{\textcolor[rgb]{.219, .713, 1}{#1}}
\newcommand{\orange}[1]{\textcolor[rgb]{1, .568, .302}{#1}}

\usepackage{multirow}

\title{Sequential Amodal Segmentation via Cumulative Occlusion Learning}

\addauthor{Jiayang Ao}{}{1} 
\addauthor{Qiuhong Ke}{}{2}
\addauthor{Krista A. Ehinger}{}{1}

\addinstitution{
 The University of Melbourne\\
 Parkville, 3010, Australia
}
\addinstitution{
    Monash University\\
    Clayton, 3800, Australia
}

\begin{document}

\maketitle

\begin{abstract}
To fully understand the 3D context of a single image, a visual system must be able to segment both the visible and occluded regions of objects, while discerning their occlusion order. Ideally, the system should be able to handle any object and not be restricted to segmenting a limited set of object classes, especially in robotic applications. Addressing this need, we introduce a diffusion model with cumulative occlusion learning designed for sequential amodal segmentation of objects with uncertain categories. This model iteratively refines the prediction using the cumulative mask strategy during diffusion, effectively capturing the uncertainty of invisible regions and adeptly reproducing the complex distribution of shapes and occlusion orders of occluded objects.
%This model considers the uncertainty of occluded objects and is adept at reproducing the complex distribution associated with their shape and occlusion order. 
It is akin to the human capability for amodal perception, i.e., to decipher the spatial ordering among objects and accurately predict complete contours for occluded objects in densely layered visual scenes. Experimental results across three amodal datasets show that our method outperforms established baselines. The code will be released upon paper acceptance.
\end{abstract}

%-------------------------------------------------------------------------
\section{Introduction}
\label{sec:intro}
Robots often encounter unfamiliar objects in ever-changing unstructured environments such as warehouses or homes~\cite{tian2023data}. These scenarios require systems capable of manipulating objects based on their complete shape and occlusion relationships rather than their visibility or category~\cite{back2022unseen, Wada2019, fu2024taylor}. However, most state-of-the-art amodal segmentation methods~\cite{ao2024amodal, Li_2023_ICCV, gao2023coarse, Tran_2022_BMVC}, which are usually constrained by the need for class-specific data, struggle to generalize to unseen objects and are susceptible to misclassification. 

Diffusion probabilistic models specialize in capturing and reproducing complex data distributions with high fidelity~\citep{ho2020denoising}, making them well-suited for generating the invisible parts of unknown objects. In contrast to traditional convolutional networks that often struggle with the complexity of occlusions \citep{ren2015faster, he2017mask}, diffusion models proficiently reconstruct objects through their iterative refinement process. This process is particularly advantageous for inferring occluded object regions, as it progressively recovers the occluded parts based on visible context and learned possible object shapes. Additionally, while current amodal segmentation methods typically overlook the uncertainty in the shape of the hidden part, diffusion models inherently sample from the learned distribution~\citep{zbinden2023stochastic, rahman2023ambiguous}, providing multiple plausible hypotheses for the occluded shape. Given these capabilities, diffusion models present a fitting approach for advancing the field of amodal segmentation.

\begin{figure}[t]
  \centering
   \includegraphics[width=0.9\linewidth]{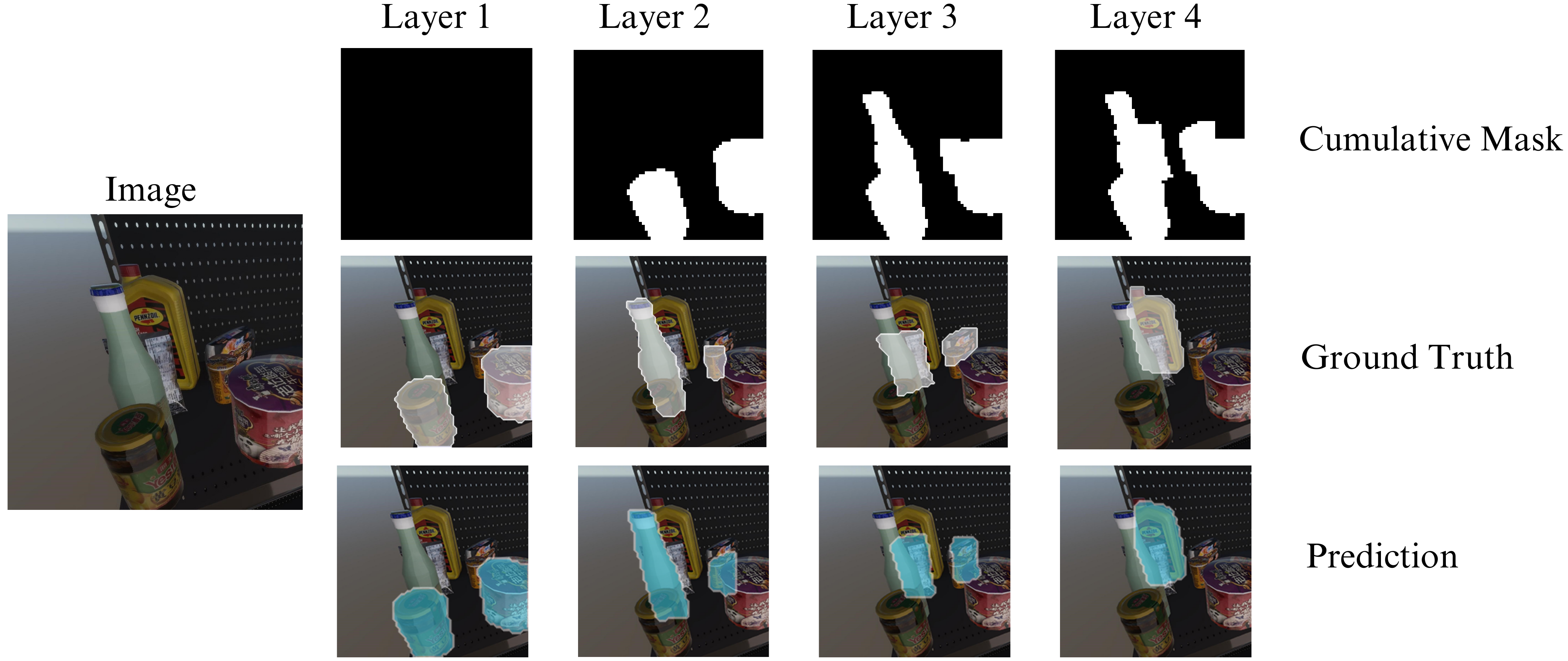}
   \caption{The cumulative mask and amodal mask predictions for an input image. Our method can generate reliable amodal masks layer by layer and allows multiple objects per layer.}
   \label{fig:demo}
\end{figure}

We introduce a novel diffusion model for sequential amodal segmentation that does not rely on object categories. Our approach transcends traditional single or dual-layer prediction limitations~\cite{qi2019amodal, ke2021deep, jiang2024} by enabling the simultaneous segmentation of unlimited object layers in an image. In addition, our framework generates multiple plausible amodal masks for each object from a single input image, contrasting with prior approaches that depend on multiple ground truths to achieve varied results~\cite{wolleb2022diffusion, rahman2023ambiguous, gao2023modeling}. Tailored to the amodal task, our method requires only a single ground truth per object during training to capture the diversity of occlusions, overcoming the limitations of existing amodal datasets that typically provide only one annotation per object and neglect the variability in invisible regions.

Our framework takes an RGB image as input and sequentially predicts the amodal masks for each object, as illustrated in Fig.~\ref{fig:demo}. The iterative refinement process of our proposed algorithm, inspired by human perception mechanisms for invisible regions~\cite{rensink1998early}, leverages preceding identified items to infer subsequent occluded items. Specifically, it employs a cumulative mask, which aggregates the masks of previously identified objects. This strategy allows the model to maintain a clear record of areas already segmented, directing its focus toward unexplored regions. By focusing the prediction effort on uncertain or occluded regions, our approach improves the accuracy and reliability of the amodal segmentation process.

We validate our approach through comprehensive ablation studies and performance benchmarking across three amodal datasets, demonstrating its superiority in handling complex sequential amodal segmentation challenges.

The main contributions of our work are: 

\begin{itemize}
\setlength{\itemsep}{6pt}
\item A new sequential amodal segmentation method capable of predicting unlimited layers of occlusion, enabling occlusion modelling in complex visual scenes. 
\item Occluded shape representation which is not based on labelled object categories, enhancing its applicability in diverse and dynamic settings.
\item A diffusion-based approach to generating amodal masks that captures the uncertainty over occluded regions, allowing for diverse segmentation outcomes. 
\end{itemize}

\section{Related Work}\label{sec:literature}

\noindent\textbf{Amodal segmentation with order perception} requires segmentation of the entire objects by including both visible and occluded regions while explicitly resolving the layer order of all objects in the image. Establishing layering of objects allows for a comprehensive understanding of the scene and the spatial relationships between objects, which is essential for tasks such as autonomous driving, robot grasping, and image manipulation~\cite{back2022unseen, lee2022instance, zheng2021visiting}. Current amodal segmentation methods mainly assess occlusion states of individual objects~\cite{qi2019amodal, follmann2019learning, sun2022bayesian, Reddy2022WALT} or between pairs~\cite{ke2021deep, yuan2021robust, back2022unseen}, but tend to ignore the global order in a complex scene, such as the relationship between independent groups. While some work~\cite{ao2024amodal, zheng2021visiting} has begun to address amodal segmentation with perceptible order, they fall short for class-agnostic applications due to design constraints on category-specific dependencies.

\textbf{Class-agnostic segmentation} aims to detect masks without relying on pre-learned category-specific knowledge. It is vital for scenarios where comprehensive labelling is resource-intensive or when encountering unseen categories~\cite{tian2023data, qi2022ssl}. However, amodal segmentation approaches usually depend on predefined class labels and thus have limited ability to handle unknown objects~\cite{Li_2023_ICCV,mohan2022amodal}. While there are a few methods which consider the class-agnostic amodal segmentation, \cite{back2022unseen} is for RGB-D images with depth data rather than RGB images, \cite{ehsani2018segan} relies on the bounding box of the object as an additional input to predict amodal masks, \cite{zhu2017semantic} treats amodal masks prediction and ordering as separate tasks thus designs the methods individually, and other requires additional inputs for prediction such as visible mask~\cite{Nguyen2021, zhan2020self}

\textbf{Segmentation with diffusion models} has recently attracted interest as its ability to capture complex and diverse structures in an image that traditional models might miss~\cite{Xu_2023_CVPR, Li_2023_diff, wu2023medsegdiff, couairon2023diffedit}. Particularly in medical imaging, diffusion models are used to generate multiple segmentation masks to simulate the diversity of annotations from different experts~\cite{wolleb2022diffusion, zbinden2023stochastic, rahman2023ambiguous, gao2023modeling}. However, these methods are designed for the visible part of images and do not adequately address the diversity of predictions required for the hidden part of objects.

In summary, our approach addresses sequential amodal segmentation with two key improvements: First, a novel segmentation technique capable of globally predicting occlusion orders, offering a comprehensive understanding of object occlusion relationships in a scene. Second, a diffusion-based model to provide diverse predictions for amodal masks, especially for the occluded portions. This model uniquely employs cumulative occlusion learning that utilises all preceding masks to provide vital spatial context, thus boosting its ability to segment occluded objects.

\section{Problem Definition}

Our goal is to amodally segment multiple overlapping objects within an image without object class labels, while determining the occlusion order of these objects. Specifically, the task requires inferring complete segmentation masks of all objects, including both the visible and occluded portions, and assigning a layering order to these segments.

For a given RGB image \( I \), the goal of our sequential amodal segmentation approach is two-fold. First, to produce a collection of amodal segmentation masks \(\{M_i\}_{i=1}^N\), where each mask \(M_i\) represents the full extent of the corresponding object \(O_i\) within the scene—this includes both visible and occluded regions. Second, to assign a layer ordering \(\{L_i\}_{i=1}^N\) to these objects based on their mutual occlusions, thereby constructing an occlusion hierarchy.

The layer variable \(L_i\) adheres to the occlusion hierarchy defined by \cite{ao2024amodal}. The bi-directional occlusion relationship \(Z(i, j)\) indicates if \(O_i\) is occluded by \(O_j\), given by:
\begin{equation}
Z(i, j) = 
\begin{cases} 
1, & \text{if object } O_i \text{ is occluded by object } O_j, \\
0, & \text{otherwise}.
\end{cases}
\end{equation}

The set \( S_i \) comprises indices of those objects occluding \(O_i\), is defined by \( S_i = \{ j | Z(i, j) = 1 \} \). Subsequently, the layer ordering \( L_i \) for each object \(O_i\) is computed based on:

\begin{equation}
L_i = 
\begin{cases} 
1, & \text{if } S_i = \emptyset, \\
1 + \max_{j \in S_i} L_j, & \text{otherwise}.
\end{cases}
\end{equation}

The ultimate goal is to derive an ordered sequence of amodal masks \( \tau = \langle M_1, \ldots, M_N \rangle \) that correctly represents the object layers in image \( I \).

\section{Methodology}
\label{sec:method}

\begin{figure*}[ht]
  \centering
   \includegraphics[width=0.96\linewidth]{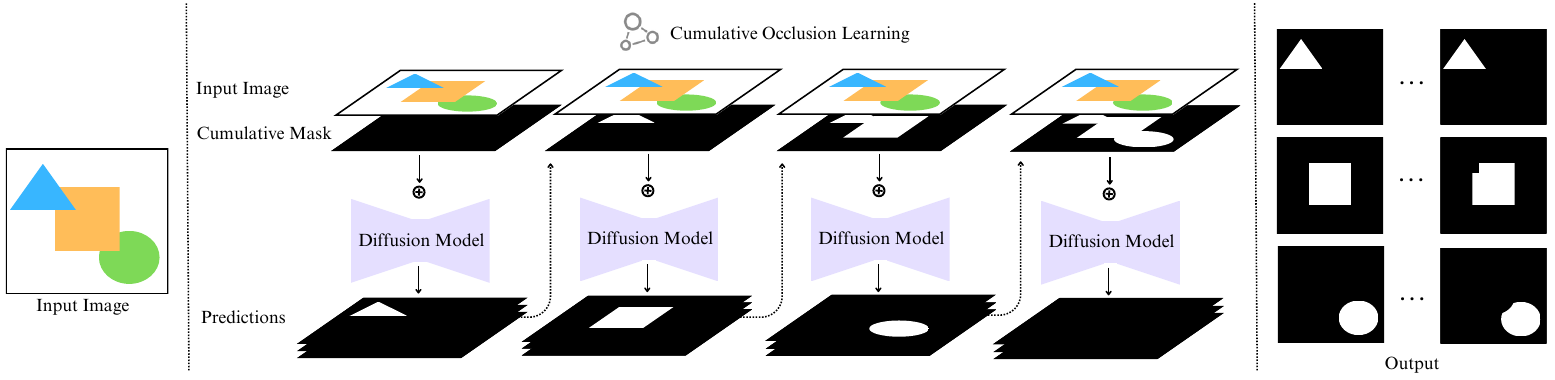}
   \caption{Architecture of our model. Our model receives an RGB image as input and predicts multiple plausible amodal masks layer-by-layer, starting with the unoccluded objects and proceeding to deeper occlusion layers. Each layer's mask synthesis receives as input the cumulative occlusion mask from previous layers, thus providing a spatial context for the diffusion process and helping the model better segment the remaining occluded objects.}
   \label{architecture}
\end{figure*}

The architecture of our proposed model is shown in Fig.~\ref{architecture}. Details on the architectural components, the cumulative guided diffusion model and the cumulative occlusion learning algorithm are discussed in Sections~\ref{sec:diffusion} and~\ref{sec:ocl}, respectively.

\subsection{Diffusion-based Framework}
\label{sec:diffusion}

Denoising diffusion probabilistic models (DDPM) are popular generative models that provide powerful frameworks for learning complex data distributions \citep{ho2020denoising}. Building on the improved DDPMs \citep{nichol2021improved}, we introduce a novel approach that extends the capabilities of diffusion models to the domain of amodal segmentation, which involves segmenting visible regions while inferring the shapes of occluded areas. This is distinct from existing diffusion models that focus primarily on visible image features, where additional understanding of occlusion structure in an image makes it a unique challenge.

\textbf{Cumulative mask.} We introduce the cumulative mask—a critical innovation that incorporates the spatial structures of objects, facilitating the understanding of both visible and occluded object parts. The cumulative mask aggregates the masks of all objects which are in front of (and potentially occluding) the current layer. Specifically, the cumulative mask for an object \( O_i \) with layer order \( L_i \) encompasses the masks of all objects with a layer order lower than \( L_i \), thereby representing the cumulative occlusion up to that layer. For each object \( O_i \) with its amodal mask \( M_i \) and layer order \( L_i \), the cumulative mask \( \text{CM}_i \) is formalized as:
\begin{equation}
    \text{CM}_i = \bigcup_{\{j | L_j < L_i\}} M_j,
\end{equation}

\noindent where \( \bigcup \) denotes the union operation, \( \text{CM}_i \) is the cumulative mask for object \( O_i \), \( M_j \) are the masks of objects with a lower layer order \( L_j \) than that of \( O_i \), reflecting the cumulative occlusion encountered up to object \( O_i \). \( \text{CM} = \varnothing \) denotes no prior occlusion and is used for the fully visible objects in \( L_1 \). 

\begin{figure}
  \centering
   \includegraphics[width=.75\linewidth]{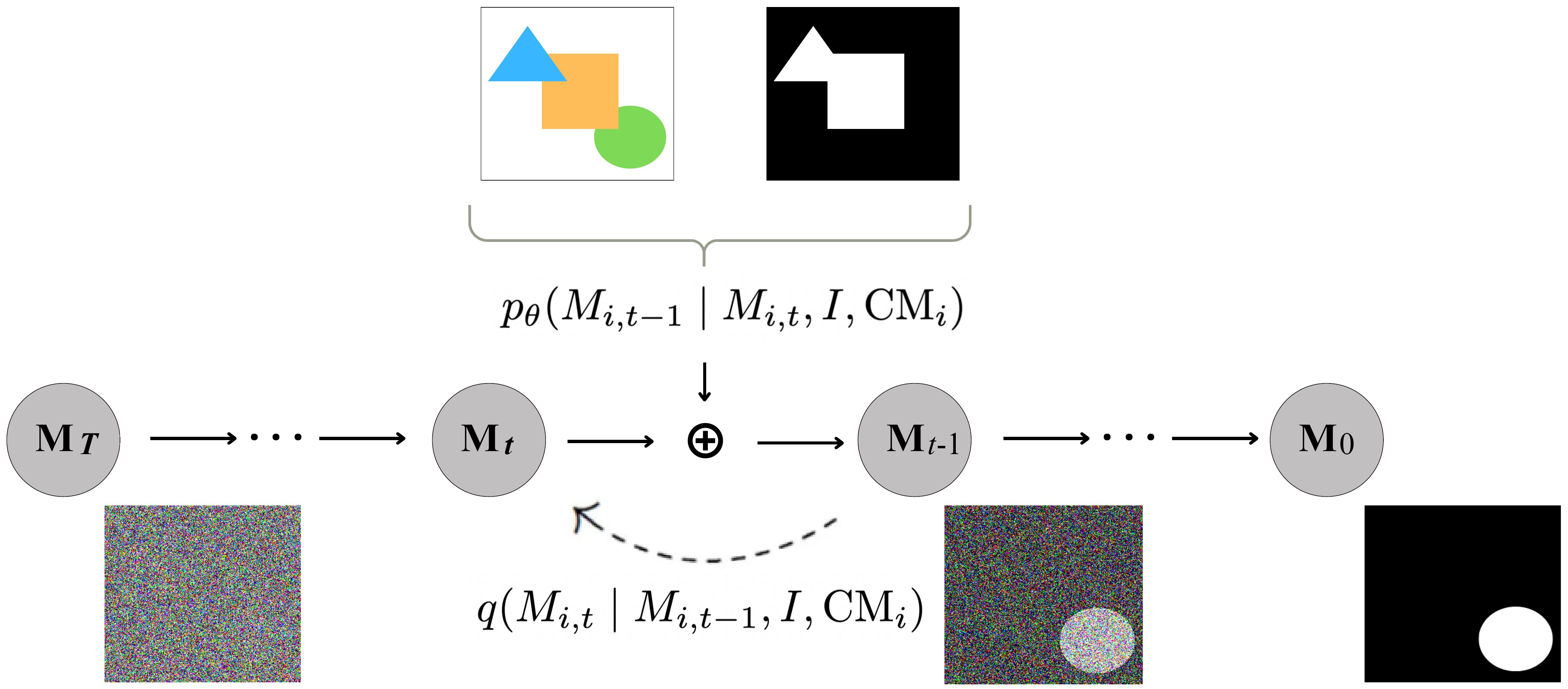}
   \caption{Cumulative guided diffusion. The diffusion process is informed by the input image and the dynamically updated cumulative mask at each depth layer. The diffusion only perturbs the amodal masks, maintaining the contextual and spatial integrity of the image and the corresponding cumulative mask unaltered.}
   \label{cplm}
\end{figure}

\textbf{Cumulative guided diffusion.} We enhance DDPMs \cite{ho2020denoising, nichol2021improved} to address the unique challenge of understanding occluded regions for amodal segmentation. The diffusion process is informed by a static representation of the input image and the cumulative mask from previous layers. The diffusion process generates an amodal mask for the current layer's objects, which is then added to the cumulative occlusion mask to generate the next layer. Fig.~\ref{cplm} illustrates the proposed cumulative guided diffusion process.

Following the standard DDPMs implementation \cite{ho2020denoising}, the diffusion process is modelled as a Markov chain. The forward process \( q \) at time \( t \) evolves from the previous step \( t-1 \) is:
\begin{equation}
    q(x_t | x_{t-1}) := \mathcal{N}(x_t; \sqrt{\alpha_t} x_{t-1}, (1-\alpha_t)\mathbf{I}),
\end{equation}
\noindent where \( x_t \) is the noisy data at \( t \), \( \alpha_t \) is the scheduler which determines the noise variance at each step, and \( \mathbf{I} \) is the identity matrix.

The reverse process, which is a learned neural network parameterized by \( \theta \), endeavours to reconstruct the original data from its noisy version, thus performing denoising:

\begin{equation}
    p_{\theta}(x_{t-1} | x_t) := \mathcal{N}(x_{t-1}; \mu_{\theta}(x_t, t), \Sigma_{\theta}(x_t, t)),
\end{equation}
\noindent where the parameters of mean and variance are \( \mu_{\theta}\) and  \(\Sigma_{\theta}\).

As proven in \citet{ho2020denoising}, \(x_{t-1}\) can be computed from \(x_{t}\):
\begin{equation}\label{xt-1}
    x_{t-1}=\frac{1}{\sqrt{\alpha_{t}}}(x_{t}-\frac{1-\alpha_{t}}{\sqrt{1-\bar{\alpha}_{t}}}\epsilon_{\theta}(x_{t},t))+\sigma_{t}\mathbf{z}
\end{equation}
\noindent where $\mathbf{z}\sim\mathcal{N}(0, \mathbf{I})$, $\bar{\alpha}_{t}:=\prod_{s=1}^{t}\alpha_{s}$, $\epsilon_{\theta}$ is a neural network function that learns noise prediction, and $\sigma_{t}$ is the standard deviation schedule.

We inform our model with the input image and its dynamically updated cumulative mask at each depth layer. This allows the model to recover the occluded objects progressively based on previously learned context. We achieve this by concatenating a given image \( I \), the cumulative mask \( \text{CM}_i \) and amodal mask \( M_{i} \) for objects in layer \( L_i \) along the channel dimension, and define:
\begin{equation}
\mathbf{X_{i}}:=I\oplus\text{CM}_i\oplus{M}_{i}
\end{equation}
\noindent The forward processing of \( q \) adds noise only to the amodal masks, keeping the input image and the corresponding cumulative mask unaltered. For a given image \( I \) and cumulative mask \( \mathrm{CM}_i \), we only add noise to the amodal mask \( M_{i} \):
\begin{equation}
    M_{i,t} = \sqrt{\bar{\alpha}_{t}}M_{i}+\sqrt{1-\bar{\alpha}_{t}}\epsilon,\quad \epsilon\sim   \mathcal{N}(0,\mathbf{I})
\end{equation}
\noindent Since we can define $\mathbf{X_{i, t}}:=I\oplus\text{CM}_i\oplus{M}_{i, t}$, Equation \ref{xt-1} is modified as,
\begin{equation}
    M_{i,t-1}=\frac{1}{\sqrt{{\alpha }_{t}}} (M_{i,t}-\frac{1-\alpha_{t}}{\sqrt{1-\overline{\alpha }_{t}}}\epsilon_{\theta }(X_{i,t},t)) + \sigma _{t}\mathbf{z}
\end{equation}
\noindent where $\mathbf{z}\sim\mathcal{N}(0, \mathbf{I})$. The reverse process aims to reconstruct the noise-free amodal mask from its noisy counterpart, effectively denoising the mask at each timestep as \( t \) decreases.

The neural network's parameters are trained to minimize the difference, measured by the Kullback-Leibler divergence, between the forward and reverse distributions across all timesteps. The loss function is expressed as:
\begin{align}
\mathcal{L}(\theta) = \mathbb{E}_{t, \mathbf{M}_i, \boldsymbol{\epsilon}} \left[ \| \boldsymbol{\epsilon} - \boldsymbol{\epsilon}_\theta(\sqrt{\bar{\alpha}_{t}}M_{i}+\sqrt{1-\bar{\alpha}_{t}}\epsilon, t) \|^2 \right],
\end{align}
\noindent where \( \boldsymbol{\epsilon} \) is the true noise, and \( \boldsymbol{\epsilon}_\theta \) is the model-predicted noise. The training process optimizes \( \theta \) by minimizing the mean squared error between the true and predicted noise, facilitating a precise recovery of the amodal mask through the reverse diffusion sequence.

During inference, the model utilizes the learned reversal mechanism to generate multiple plausible amodal masks by sampling from a standard Gaussian distribution and conditioning on each object's unique context:
\begin{equation}
    M_{gen,i}^{(k)} = f_{\theta}(\mathcal{N}(0, \mathbf{I}), I, \text{CM}_i), \quad k = 1, \dots, K,
\end{equation}

\noindent where \( f_{\theta} \) represents the trained generative function of the model, and \( M_{gen,i}^{(k)} \) is the \( k \)-th generated amodal mask prediction for the object \( O_i \). This process allows the generation of multiple plausible occlusion masks for each object layer.

Through the specialized design of cumulative guided diffusion, our framework is able to address the intricacies of the sequential amodal segmentation task, uncovering the full scope of objects in complex, multi-layered visual scenes.

\subsection{Cumulative Occlusion Learning}
\label{sec:ocl}

Lack of spatial contextual awareness of surrounding objects in amodal segmentation can yield inaccurate or incomplete scene interpretations. To address this, we propose the cumulative occlusion learning algorithm, which employs a hierarchical procedure that learns to predict amodal masks in an order-aware manner. It operates by accumulating visual information, where the history of observed data (previous segmentation masks) influences the perception of the current data (the current object to segment). This strategy is akin to human perception, where the understanding of a scene is constructed incrementally as each object is identified and its spatial relation to others is established.

\textbf{Training.} We initiate with an empty cumulative mask (\(\text{CM}_0\)) and an image \(I\) with \(N\) layers. The model proceeds iteratively, predicting the amodal mask \(\hat{M}_i\) for each layer while updating the cumulative mask using ground truth amodal masks to ensure the accuracy of the spatial context during training. Note that the diffusion is applied solely to the amodal mask predictions, while the image \( I \) and the cumulative mask \( \text{CM} \) remain intact. This cumulative strategy enhances accuracy by incorporating occlusion context into each layer in the learning process, enhancing the model's spatial understanding. Alg.~\ref{training} shows the complete training process. Notably, we introduce a predictive step for a layer \(N+1\), which trains the model to expect a blank mask after all object layers have been identified and segmented. This ensures that the model learns to identify the last layer with any partially-visible objects and does not continue to hallucinate fully-occluded objects behind these. 

\begin{algorithm}[t]
\scriptsize
\caption{Training Algorithm for cumulative occlusion learning}
\label{training}
\begin{algorithmic}
\State \textbf{Input:} Image \( I \) with number of \( N \) layers
\State \textbf{Output:} Ordered sequence of amodal masks \( \tau = \langle \hat{M}_1, \hat{M}_2, \ldots, \hat{M}_N \rangle \)
\State Initialize \( \text{CM}_0 \) to a blank mask
\State Initialize the ordered sequence \( \tau \) as an empty list
\For{\( i = 1 \) to \( N\)}
    \State Input to model: \( I \), \( \text{CM}_{i-1} \)
    \State Predict amodal mask \( \hat{M}_i \)  for objects in layer \( L_i \)
    \State Update \( \text{CM}_{i} \leftarrow \text{CM}_{i-1} \cup {M}_i \) (Ground Truth)
    \State Append \(\hat{M}_i \) to the sequence \( \tau \)
\EndFor
\State Perform a final prediction \( \hat{M}_{N+1} \) with \( I \) and \( \text{CM}_N \)
\State \textbf{assert} \( \hat{M}_{N+1} \) is a blank mask
\State \textbf{return} \( \tau \)
\end{algorithmic}
\end{algorithm}

\textbf{Inference.} 
Different from training, the inference phase needs to operate without available ground truth. Thus, it selects the most probable amodal mask from multiple predictions generated by the diffusion model to update the cumulative mask. Inference commences with an image \( I \) and aims to reconstruct an ordered sequence of amodal masks by layer. For each layer, a set of \( K \) diffusion-generated amodal mask predictions are evaluated to select the most representative amodal mask \(\hat{M}_i\) for that layer. The selection criterion is based on the minimum absolute difference from each mask to the mean of non-null predictions, while ensuring spatial continuity between consecutive layers. The selected mask is then utilized to update the cumulative mask for subsequent layers' predictions. The process continues iteratively for an image \(I\) until a stopping criterion is met. The stopping criteria are established to avoid over-generation of invalid predictions when (1) reaching the maximum number of layers, or (2) all predicted masks are empty or the predicted object pixels of the selected mask are below a threshold area. 
%The stopping criteria are established to ensure computational efficiency and algorithmic reliability, avoiding the over-generation of invalid predictions. 
Alg.~\ref{inference} shows the complete inference process, where the stopping criteria \( N_{max} \) and \( Area_{min} \) are determined by the maximum number of layers and the minimum object area present in the corresponding training data, respectively. 

\begin{algorithm}
\scriptsize
\caption{Inference Algorithm for cumulative occlusion learning}
\label{inference}
\begin{algorithmic}
\State \textbf{Input:} Image \( I \), Maximum number of layers \( N_{max} \), Minimum object pixel area \( Area_{min} \)
\State \textbf{Output:} Ordered sequence of amodal masks \( \tau = \langle \hat{M}_1, \hat{M}_2, \ldots \rangle \)
\State Initialize \( \text{CM}_0 \) to a blank mask
\State Initialize the ordered sequence \( \tau \) as an empty list; Initialize \( i = 1 \)
\While{\(i \leq N_{max}\)}
    \State Generate \( K \) mask predictions \( \{ \hat{M}_i^1, \hat{M}_i^2, \ldots, \hat{M}_i^K \} \)
    \State Compute mean map \( \overline{M}_i \) from non-null \( \hat{M}_i^j \)
    \State Select \( \hat{M}_i \) with minimum \( ||\hat{M}_{i}^{k} - \overline{M}_i|| \)
    \State Enforce spatial integrity: if \( \hat{M}_i \cap \hat{M}_{i-1} = \varnothing \), reassign \( \hat{M}_i \) to the same layer as \( \hat{M}_{i-1} \)
    \If{\( \hat{M}_i \) is null or \( \hat{M}_i.area < Area_{min} \)} Break
    \EndIf
    \State Update \( \text{CM}_i \leftarrow \text{CM}_{i-1} \cup \hat{M}_i \)
    \State Append \( \hat{M}_i \) to \( \tau \)
    \State \( i \leftarrow i + 1 \)
\EndWhile
\State \textbf{return} \( \tau \)
\end{algorithmic}
\end{algorithm}

\textbf{Strategies for using ground truth or predicted cumulative mask.} Our model leverages the ground truth cumulative mask as input during training, while inference uses the predicted masks from previous layers to build the cumulative mask. A common idea is to utilize the predicted cumulative mask in training, mirroring the inference setup. However, this complicates the early stages of training, when all of the predicted masks (and thus the cumulative mask) are similar to random noise. We conducted experiments in which we introduced controlled noise into the cumulative mask during training, to simulate the types of errors which occur during inference, but the results showed that this did not noticeably change the trained model's performance (see Sec.~\ref{noise}). Therefore, the model presented here uses the ground truth cumulative mask during training.

In summary, cumulative occlusion learning allows the network to learn a robust internal representation of class-agnostic amodal object shape through occlusion and learns to recognize the depth layer ordering of objects in scenes. This approach means the model can any arbitrary number of layers of occlusions, because it automatically learns to recognise when all visible objects have been segmented. Moreover, by preserving the input image and cumulative mask unaltered during the diffusion perturbations, our model maintains the fidelity of the contextual information, which is crucial for generating accurate amodal predictions.

\section{Experiments and Discussions}
\label{sec:experiment}

\subsection{Datasets}
We focus on amodal datasets highly relevant to robotics applications. Intra-AFruit, ACOM and MUVA~\cite{ao2024amodal, Li_2023_ICCV} include objects such as fruits, vegetables, groceries, and everyday products, effectively simulate the kind of visual clutter and occlusion challenges encountered in industrial robotics, making them ideal for our study. We enhanced these three datasets tailored for novel sequential amodal segmentation tasks, with layer structure annotations and class-agnostic masks. The training and test images in these datasets are sourced directly from the corresponding partitions of the original dataset. All images have been downsampled to a resolution of 64 × 64 pixels for computational efficiency. To eliminate indistinguishable or misleading ground truth data, we excluded images with post-downsampling visible object areas under 10 pixels.

\textbf{Intra-AFruit~\citep{ao2024amodal}} dataset contains ten classes of fruits and vegetables. We limited the original test set to a random subset of 3,000 images to enhance experimental efficiency. The reprocessed dataset includes 187,204 training and 3,000 test images, with each image potentially containing up to five layers.

\textbf{ACOM~\citep{ao2024amodal}} dataset contains ten classes of common objects with synthetically generated annotations. The reprocessed dataset includes 9,378 training and 2,355 test images with up to five layers.

\textbf{MUVA~\citep{Li_2023_ICCV}} dataset contains twenty categories of supermarket items. To avoid compression distortion of non-square images, we cropped square images using the shortest edge and aligned the crop to the leftmost or centre, which follows object distribution rules to preserve more objects. The reprocessed dataset includes 5,582 training and 1,722 test images with up to seven layers.

\subsection{Implementation Details}
We set the timestep T=1,000 with a linear noise schedule for all the diffusion models. The models were trained using the AdamW optimizer~\citep{loshchilov2017decoupled} at a learning rate of 0.0001 and a batch size of 256. The other hyperparameters of the diffusion models follow the implementation in~\citep{nichol2021improved}. All experiments were implemented using the PyTorch framework and trained/tested on one A100 GPU. 

\textbf{Evaluation metrics.} The performance of class-agnostic segmentation is generally measured by comparing predicted masks with ground truth annotations~\citep{cheng2020cascadepsp, siam2021video, radwan2023distilling}. We adopted two commonly used metrics: intersection over union (IOU) and average precision (AP).

\subsection{Architecture Analysis}

\begin{figure}[h]
\begin{tabular}{ccc}
\bmvaHangBox{\fbox{\parbox{5.8cm}{~\\[2mm]
\rule{0pt}{1ex}\includegraphics[width=5.8cm]{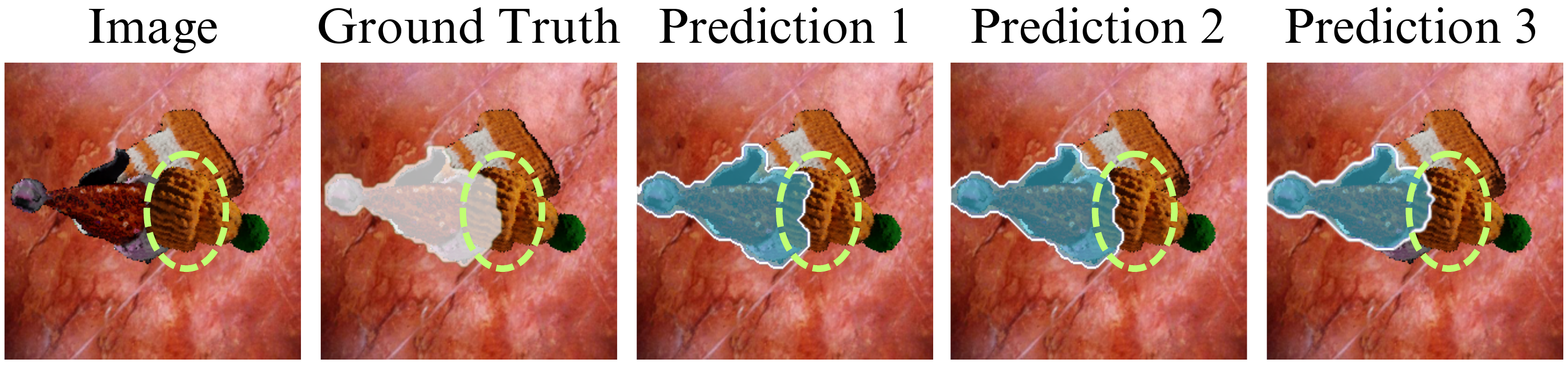}\\}}}&
\bmvaHangBox{\fbox{\includegraphics[width=5.8cm]{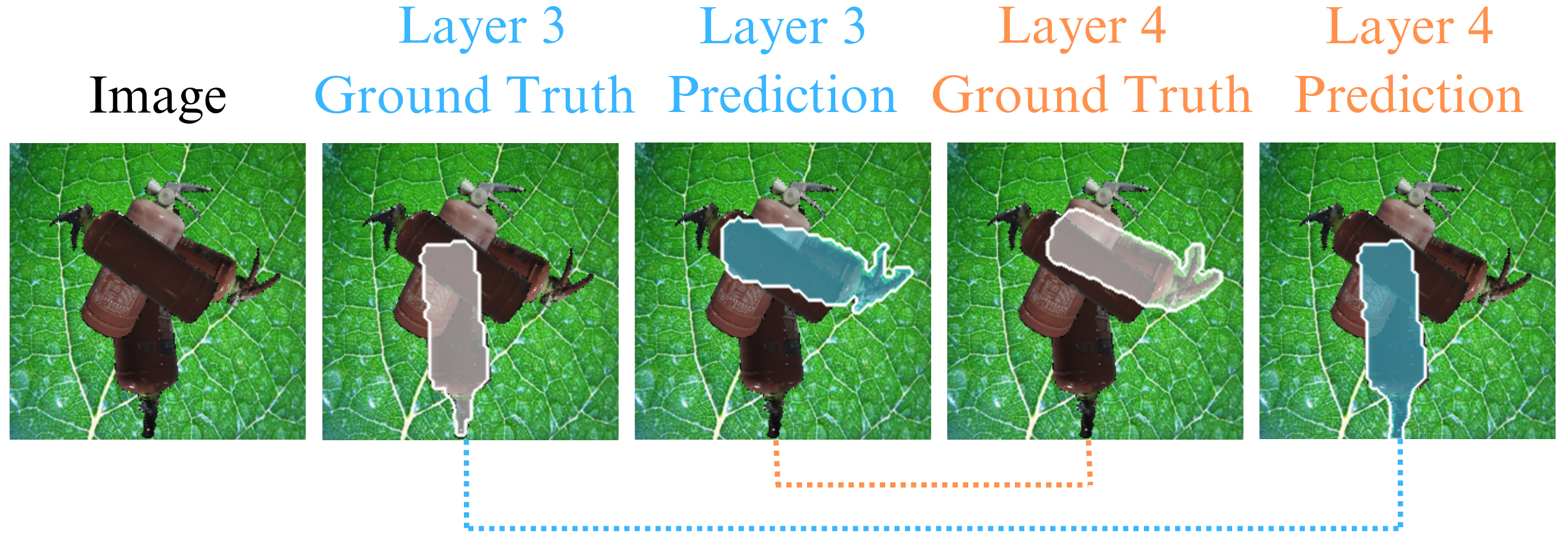}}}\\
(a)&(b)
\end{tabular}
\caption{(a) Our approach considers the diversity of possible amodal masks, especially for occluded regions (indicated by dashed circles). (b) Example of misjudgement of the order of occluded objects in adjacent layers. \blue{Layer 3's prediction} reflects \orange{Layer 4's ground truth} and vice versa. This can also be a challenge for human perception.}
\label{fig:com}
\end{figure}

\textbf{Number of generated amodal masks.} Our proposed method enables the generation of multiple amodal masks for each object, thus enabling the capture of uncertainty and allowing for the diversity of reasonable configurations of the occluded parts without the need for diverse training annotations for each image (see Fig.~\ref{fig:com} (a)). This is particularly useful for amodal tasks considering occluded areas, where manual annotation is very expensive and synthetic images often provide only the sole ground truth. 

\begin{table}[h]
\scriptsize
\begin{center}
\begin{tabular}{c|cc|cc|cc|cc|cc}
\toprule
Metric   & \multicolumn{2}{c|}{Layer 1} & \multicolumn{2}{c|}{Layer 2} & \multicolumn{2}{c|}{Layer 3} & \multicolumn{2}{c|}{Layer 4} & \multicolumn{2}{c}{Layer 5} \\
\midrule
Ensemble & IOU           & AP           & IOU          & AP           & IOU          & AP           & IOU          & AP           & IOU         & AP           \\
\hline
k=3      & \textbf{57.1} & \textbf{57.8} & \textbf{44.8} & \textbf{45.4} & 28.8         & 30.0         & 12.2         & 14.2         & 1.9         & 3.6          \\
k=5      & 56.7          & 57.5          & 44.3         & 44.9          & 28.8         & 29.7         & 12.7         & \textbf{14.3} & 2.3         & \textbf{3.7} \\
k=7      & 56.8          & 57.5          & 44.7         & 45.4          & 29.4         & 30.0         & 12.6         & 14.1         & 2.6         & 3.6          \\
k=9      & 56.9          & 57.7          & 44.4         & 45.1          & \textbf{29.5} & \textbf{30.2} & \textbf{12.9} & 14.2        & 2.4         & 3.7          \\
\bottomrule
\end{tabular}
\end{center}
\caption{Ablation study for generating different numbers of masks during inference.} 
\label{tab:numberofmask}
\end{table}

While an arbitrary number of masks could be generated, we need to set a reasonable number for inference. Tab.~\ref{tab:numberofmask} shows the performance of generating different numbers of masks for each layer during inference on the ACOM dataset, where the IOU and AP do not vary much, but the computation increases dramatically with more masks. Considering the computational efficiency, we generated 3 masks per layer in subsequent experiments.

\textbf{Selection of cumulative mask.}
The inference process could give multiple predictions for each layer, so there might be two options to update the cumulative mask for a given layer: (1) use one most plausible prediction for that layer. Here, we choose the prediction with the minimum absolute difference from the mean of all predictions as the one. (2) use the mean of all predictions for that layer to form a mean mask. While the mean mask more explicitly takes into account all predictions, the risk is that when a prediction incorrectly gives an object that does not belong in that layer, the mean mask reacts to that as well. For example, a previous prediction showing an object in the next layer may cause the next prediction to ignore that object, because the object is already included in the given mean mask. 

Therefore, in the inference process, the cumulative mask employs the most representative amodal mask (with the minimum absolute difference from the mean mask) rather than directly using the mean mask of all predictions for that layer. This avoids confusion due to the simultaneous prediction of objects in different layers. Tab.~\ref{tab:meanmap} shows the superiority of our mask selection method over using the mean mask for occluded layers on ACOM dataset.

\begin{table}[!h]
\scriptsize
\centering
\begin{tabular}{c|ccccc}
\toprule
Choice of       & L1 & L2 & L3 & L4 & L5 \\ \cline{2-6} 
Cumulative Mask & \multicolumn{5}{c}{AP} \\ \hline
Mean mask & 57.7 & 43.1 & 27.9 & 10.4 & 2.8 \\
Select mask & \begin{tabular}[c]{@{}c@{}}57.8\\ (+0.1)\end{tabular} & \begin{tabular}[c]{@{}c@{}}45.4\\ (\textbf{+2.3})\end{tabular} & \begin{tabular}[c]{@{}c@{}}30.0\\ (\textbf{+2.1})\end{tabular} & \begin{tabular}[c]{@{}c@{}}14.2\\ (\textbf{+3.8})\end{tabular} & \begin{tabular}[c]{@{}c@{}}3.6\\ (\textbf{+0.8})\end{tabular} \\\bottomrule
\end{tabular}
\vspace{+0.5cm}
\caption{For predicting occluded objects (Layer $L$\textgreater1), the mask we selected is more suitable for constructing cumulative masks than using the mean mask directly.} 
\label{tab:meanmap}
\end{table}

\textbf{Failure analysis.} A common challenge arises from errors in sequential prediction, particularly determining which of two objects is in front of the other when the overlapping region is occluded by a third object. This may lead to objects being predicted in incorrect layers, as illustrated in Fig.~\ref{fig:com} (b). Synthetic images can amplify this challenge due to fewer spatial cues (such as height in the image plane or scene semantics) to disambiguate occluded object order. Our cumulative occlusion learning mitigates the impact of these errors by considering the cumulative mask for all preceding layers. We demonstrate the robustness of our method to such failures through noise introduction experiments in the next section.

\subsection{Noise Introduction Experiment in Cumulative Mask}
\label{noise}
Our model leverages the ground truth cumulative mask as input during training, while inference uses the predicted masks from previous layers to build the cumulative mask, as described in Sec.~\ref{sec:ocl}. A common idea is to utilize the predicted cumulative mask in training, mirroring the inference setup. However, this complicates the early stages of training, when all of the predicted masks (and thus the cumulative mask) are similar to random noise. 

To bridge the gap between training and inference, we conducted experiments in which we introduced controlled noise into the cumulative mask during training, to simulate the types of errors which occur during inference. The experiment was designed to mimic common types of inference errors, such as continuous prediction errors due to layer dependencies or over-segmentation due to boundary ambiguity. This was achieved by selectively omitting instances from a random layer in the cumulative mask while keeping the input RGB image and the prediction mask unchanged. 

These experiments also simulate and seek to understand the impact of sequential prediction errors on the model's performance. By introducing noise into the cumulative mask during training, we effectively create scenarios where the model must handle instances segmented into the wrong layer, as happens when the model makes sequential prediction errors. 

Specifically, instances from a randomly chosen layer (excluding the fully visible layer) are excluded from the cumulative mask. Mathematically, selecting a random layer index \( i_{\text{rand}} \) from [2, n], the perturbed version of the cumulative mask, denoted as \( P \), is derived by:

\begin{equation}
    P = CM - M_{i_{\text{rand}}}
\end{equation}

Where \( CM \) is the original cumulative mask, and \( M_i \) is the ground truth mask of the \( i^{th} \) layer instance (\( i \in [2, n] \)). The subtraction here is a pixel-wise binary operation. During training, the model will replace \( CM \) with \( P \) as input at a specified noise level ratio.

\begin{table}[h]
\centering
\footnotesize
\begin{tabular}{c|ccccc|ccccc}
\toprule
Noise & 0\% & 5\%  & 10\% & 15\% & 20\% & 0\% & 5\%  & 10\% & 15\% & 20\% \\ \hline
Layer & \multicolumn{5}{c|}{AP}& \multicolumn{5}{c}{IOU}\\ \hline
1     & \textbf{57.8} & 51.7 & 56.6 & 56.0 & 57.6& \textbf{57.1} & 50.3 & 55.8         & 55.3 & 56.9 \\
2     & \textbf{45.4} & 37.5 & 44.1 & 40.2 & 40.3& \textbf{44.8} & 35.5 & 43.2         & 38.8 & 39.2 \\
3     & \textbf{30.0} & 24.6 & 28.0 & 24.9 & 23.5& \textbf{28.8} & 21.9 & 26.8         & 22.4 & 20.8 \\
4     & \textbf{14.2} & 10.7 & 12.1 & 10.3 & 9.2& \textbf{12.2} & 7.9  & 10.3         & 8.0  & 6.5  \\
5     & \textbf{3.6}  & 3.3  & 3.4  & 3.2  & 2.9& 1.9           & 1.9  & \textbf{2.2} & 1.7  & 1.0  \\ 
\bottomrule
\end{tabular}
\vspace{+0.5cm}
\caption{Comparison at different noise levels, evaluated with AP and IOU. Noise-free training results in the highest AP across the layers, and the highest IOU for the first four layers and the second highest for the fifth layer.} 
\label{tab:noiseap}
\end{table}

Tab.~\ref{tab:noiseap} illustrates the model's performance in terms of AP and IOU across different layers and noise levels. It was observed that the highest AP was achieved with 0\% noise for all layers. Similar to AP, the IOU results also showed that the highest performance was generally observed with 0\% noise, except for the 5th layer, where a slight increase was noted at 10\% noise level. Overall, this suggests that adding noise in training has very limited benefit. On the contrary, training without noise achieves the best performance in terms of AP or IOU in the vast majority of cases.

The results of the experiment provide insight into the model's robustness to errors in the sequential segmentation process and validate the effectiveness of our cumulative occlusion learning approach. By focusing on the cumulative mask for all preceding layers, our approach avoids the cascading effects of sequential prediction errors, ensuring more reliable performance even in complex occlusion scenarios. 

Despite the theoretical appeal of mimicking inference conditions during training, the results indicate that using ground truth cumulative masks remains the more effective approach. This strategy consistently yielded superior results across most metrics and layers, showing its suitability to our model training process. Based on these findings, our training strategy uses the ground truth cumulative masks.

\subsection{Comparisons with Other Methods}

%Traditional diffusion models focus on visible image segments and do not address the segmentation of occluded object parts. While these models can learn the distribution of object masks, they still lack the ability for layer prediction. 
We benchmark against DIS~\cite{wolleb2022diffusion}, a leading diffusion-based segmentation method. For comparison, we trained distinct DIS models for each layer under the same iterations and evaluated the segmentation results separately for each layer. Tab.~\ref{tab:dis} comprehensively compares our method and the improved DIS across different layers on three amodal datasets. The performance of the MUVA dataset after five layers is omitted because the performance of both models approaches zero. The superiority of our method is particularly evident in deeper layers, where our method maintains reasonable performances, whereas DIS shows a marked decline, especially in the MUVA dataset. These results highlight the robustness of cumulative occlusion learning in handling layered occlusions across various datasets, particularly in more complex scenarios involving multiple layers of object occlusion.

\begin{table}
\begin{center}
\scriptsize
\begin{tabular}{cc|ccccc}
\toprule
& Layer & 1 & 2 & 3 & 4 & 5 \\
\midrule
Dataset & Method  & IOU / AP & IOU / AP & IOU / AP & IOU / AP & IOU / AP \\
\midrule
\multirow{2}{*}{Intra-AFruit} & DIS & 89.5 / 90.7 & 81.6 / 82.6 & 52.4 / 52.6 & 9.8 / 12.4 & 0.5 / 2.0 \\
                              & Ours & \textbf{94.3} / \textbf{94.7} & \textbf{87.4} / \textbf{88.2} & \textbf{76.2} / \textbf{77.3} & \textbf{26.7} / \textbf{27.6} & \textbf{7.2} / \textbf{7.4} \\
\midrule
\multirow{2}{*}{ACOM} & DIS & 31.6 / 34.8 & 26.6 / 28.7 & 1.6 / 10.2 & 0.2 / 6.0 & 0.1 / 2.5 \\
                      & Ours & \textbf{57.1} / \textbf{57.8} & \textbf{44.8} / \textbf{45.4} & \textbf{28.8} / \textbf{30.0} & \textbf{12.2} / \textbf{14.2} & \textbf{1.9} / \textbf{3.6} \\
\midrule
\multirow{2}{*}{MUVA} & DIS & 68.2 / 71.5 & 19.3 / 27.3 & 0.1 / 8.6 & 0.2 / 3.4 & ~0 / 0.5 \\
                      & Ours & \textbf{77.0} / \textbf{79.3} & \textbf{48.7} / \textbf{51.2} & \textbf{25.4} / \textbf{27.8} & \textbf{8.5} / \textbf{9.9} & \textbf{1.0} / \textbf{1.1} \\
\bottomrule
\end{tabular}
\end{center}
\caption{Comparison with a diffusion-based segmentation model~\cite{wolleb2022diffusion} without cumulative occlusion learning. Our method exhibits great improvement in complex, deeper-layer scenes.} 
\label{tab:dis}
\end{table}

Due to the lack of class-agnostic amodal segmentation methods with layer perception, we compare against category-specific methods like PLIn for amodal segmentation with occlusion layer prediction \cite{ao2024amodal}, AISFormer for amodal segmentation without layer perception \cite{Tran_2022_BMVC}, and PointRend for modal segmentation \cite{kirillov2020pointrend}. We trained these comparison models using category-labelled amodal masks to meet their requirement for category-specific learning, while our model is trained on data without category labels. For evaluation, we ignore category label accuracy for the comparison models, reporting only segmentation accuracy. 

\begin{table}
\begin{center}
\scriptsize
\begin{tabular}{ccc|cc|cc|cc}
\toprule
\multicolumn{3}{c|}{Dataset}& \multicolumn{2}{c|}{Intra-AFruit} & \multicolumn{2}{c|}{ACOM}& \multicolumn{2}{c}{MUVA}
\\ \hline
\begin{tabular}[c]{@{}c@{}}Method \end{tabular} & \begin{tabular}[c]{@{}c@{}}Supervision \end{tabular} & \begin{tabular}[c]{@{}c@{}}Framework \end{tabular} &
\begin{tabular}[c]{@{}c@{}}AP w/ \\ Layer\end{tabular} & \begin{tabular}[c]{@{}c@{}}AP w/o \\ Layer\end{tabular} & 
\begin{tabular}[c]{@{}c@{}}AP w/ \\ Layer\end{tabular} & \begin{tabular}[c]{@{}c@{}}AP w/o \\ Layer\end{tabular} & \begin{tabular}[c]{@{}c@{}}AP w/ \\ Layer\end{tabular} & \begin{tabular}[c]{@{}c@{}}AP w/o \\ Layer\end{tabular} \\ \hline
PointRend&Supervised&	CNN-based& N/A & 70.9& N/A & 22.0 & N/A & 38.9      \\ 
AISFormer&Supervised&Transformer-based &  N/A & 70.4 & N/A & 34.9 & N/A & 49.7      \\
PLIn&Weakly supervised	&	CNN-based& 42.2 & 78.9 & 3.9 & 17.0 & 16.3& 47.3 \\
Ours&Supervised&Diffusion-based& \textbf{84.6}& \textbf{92.6}& \textbf{45.4}& \textbf{65.5}& \textbf{53.1}& \textbf{55.7}\\\bottomrule
\end{tabular}
\end{center}
\caption{Comparison with category-specific segmentation models. PointRend~\cite{kirillov2020pointrend}, AISFormer~\cite{Tran_2022_BMVC} and PLIn~\cite{ao2024amodal} are trained on category-specific data, whereas our models are trained using class-agnostic data. We evaluate the models by focusing solely on the segmentation quality, disregarding any category information.} 
\label{tab:specific}
\end{table}

\begin{figure}
  \centering
   \includegraphics[width=\linewidth]{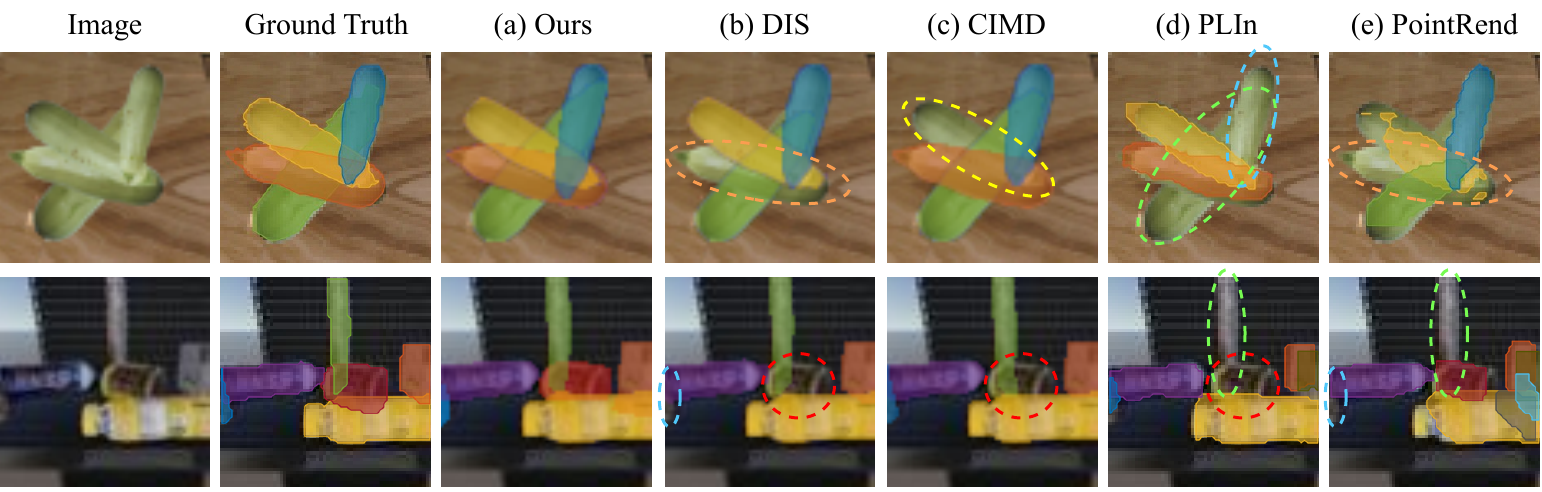}
   \caption{Comparison of predictions on Intra-AFruit (top) and MUVA (bottom) test image by (b) DIS~\cite{wolleb2022diffusion} (c) CIMD~\cite{rahman2023ambiguous} (d) PLIn~\cite{ao2024amodal} (e) PointRend~\cite{kirillov2020pointrend} and (a) ours, where (b) and (c) are diffusion-based methods. Dashed circles indicate objects that missed being predicted. Others fail to segment objects or provide less plausible amodal masks compared to ours.}
   \label{fig:prediction}
\end{figure}

We present the AP results considering two scenarios in Tab.~\ref{tab:specific}: with layer prediction, where segmentation precision is contingent on correct layer assignment, and without layer prediction, where segmentation is recognized irrespective of layer placement. Despite being trained on class-agnostic data, our method surpasses category-specific models trained on category-labelled data. Furthermore, Fig.~\ref{fig:prediction} visually demonstrates our method's superiority in amodal mask segmentation. Our approach provides plausible masks even for heavily-occluded objects, showcasing its enhanced segmentation capability in complex scenes involving multiple layers of object occlusion.

\begin{figure}
  \centering
   \includegraphics[width=\linewidth]{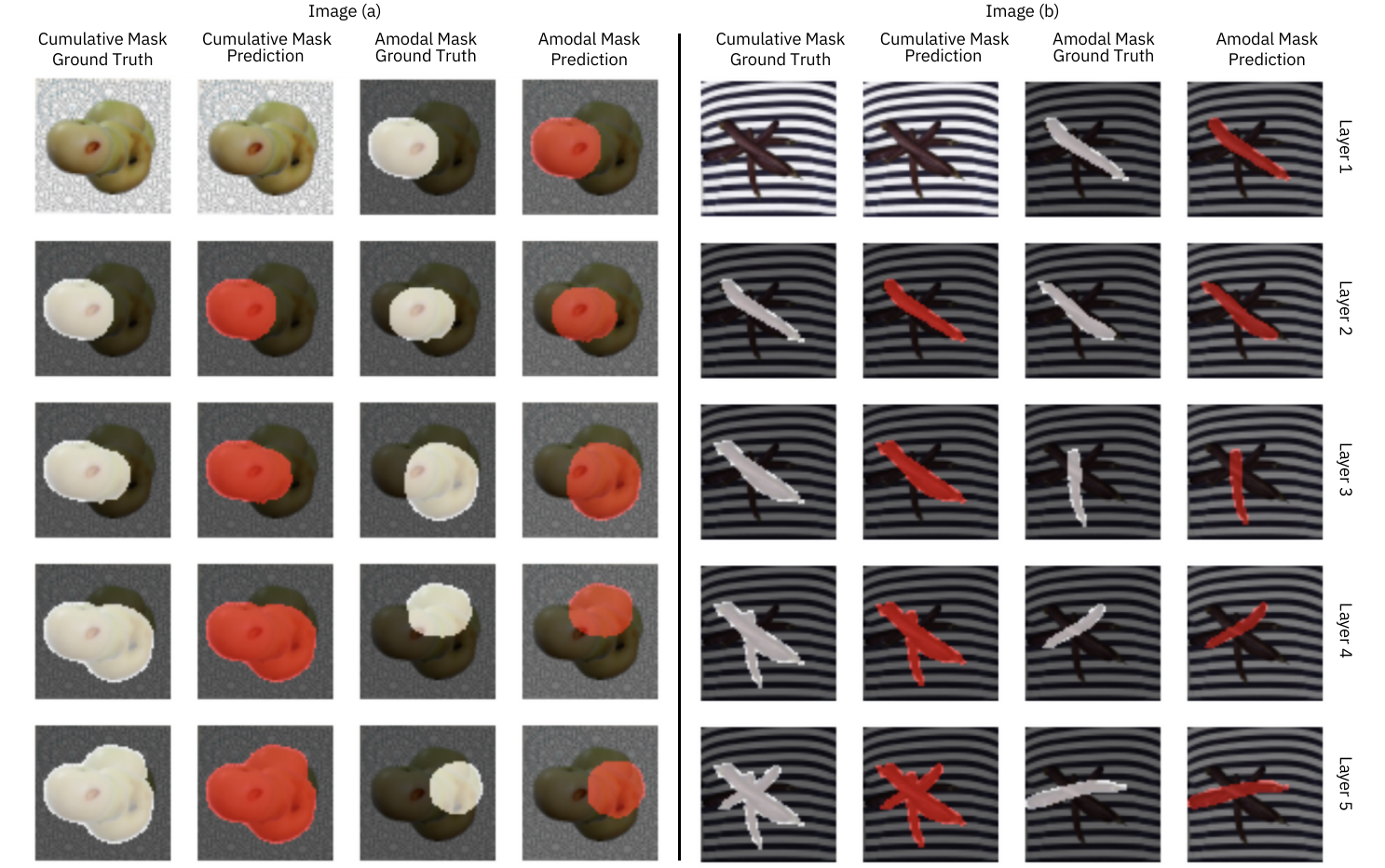}
   \caption{Visualisation of the prediction of our model on the Intra-AFruit~\cite{ao2024amodal} test set. Each layer's amodal mask synthesis receives the cumulative mask of the previous layers as input, thus providing a spatial context for the prediction and helping to segment the remaining occluded objects better. We can see that our model can predict amodal masks and occlusion layers well for multiple objects in a given image.}
   \label{fig:afruit}
\end{figure}

\begin{figure} 
  \centering
   \includegraphics[width=\linewidth]{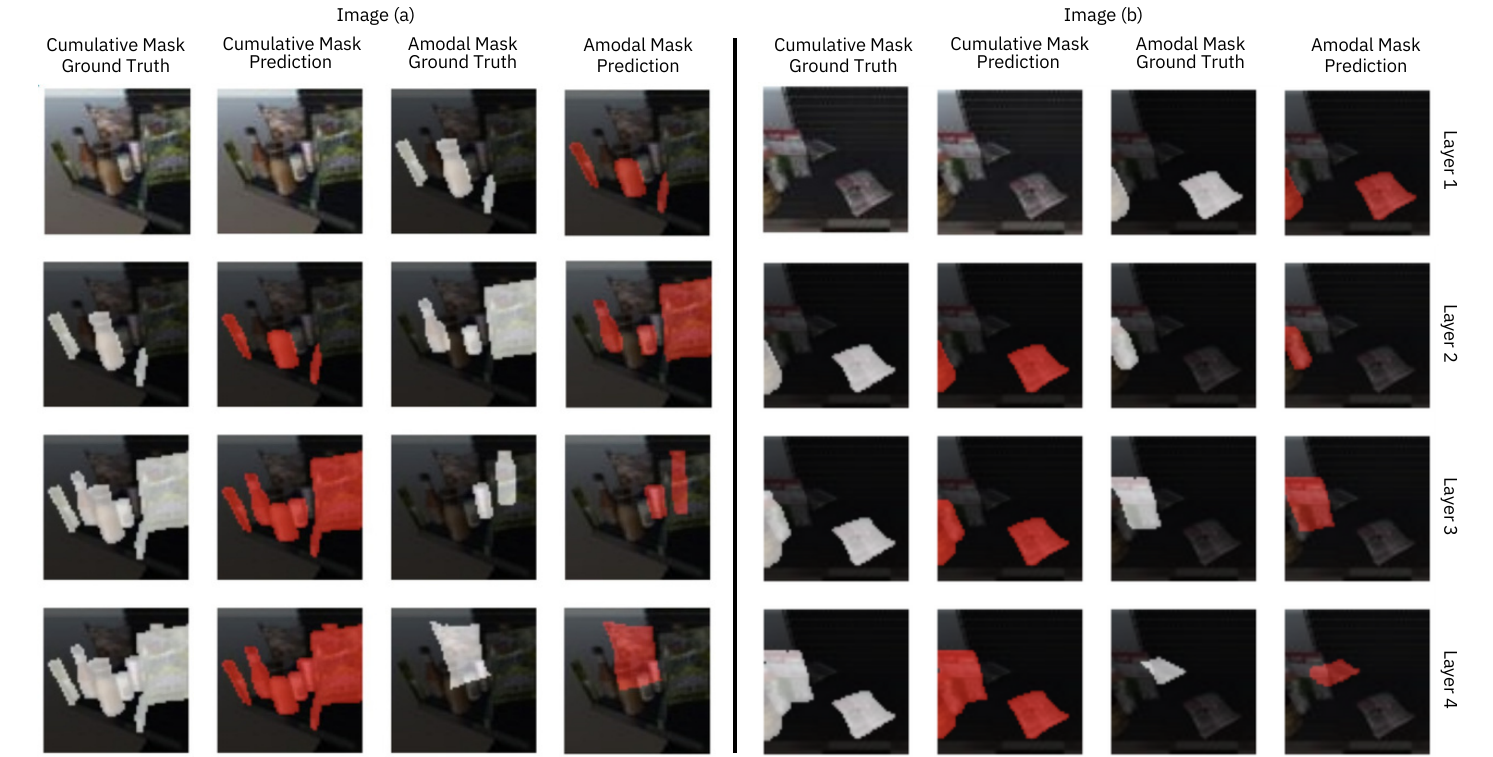}
   \caption{Visualisation of the prediction of our model on the MUVA~\cite{Li_2023_ICCV} test set.}
   \label{fig:muva}
\end{figure}

\begin{figure}
  \centering
   \includegraphics[width=\linewidth]{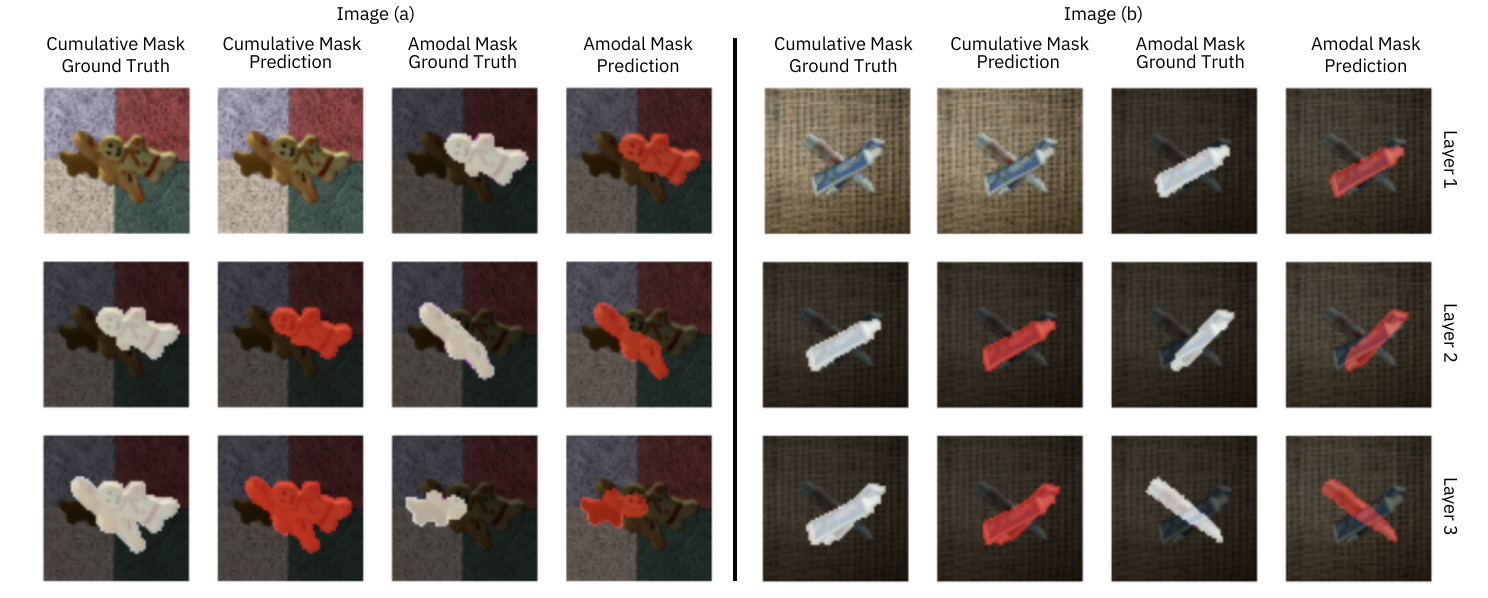}
   \caption{Visualisation of the prediction of our model on the ACOM~\cite{ao2024amodal} test set.}
   \label{fig:acom}
\end{figure}

We provide more visualisations of our model's predictions for the Intra-AFruit~\cite{ao2024amodal}, MUVA~\cite{Li_2023_ICCV} (Fig.~\ref{fig:muva}),  (Fig.~\ref{fig:afruit}) and ACOM~\cite{ao2024amodal} (Fig.~\ref{fig:acom}) test sets. As we can see from the figures, our model performs robustly with different objects and different levels of occlusion.

\section{Conclusion}
\label{sec:conclude}

The task of sequential amodal segmentation is essential for understanding complex visual scenes where objects are frequently occluded. Our proposed method, leveraging cumulative occlusion learning with mask generation based on diffusion models, allows robust occlusion perception and amodal object segmentation over unknown object classes and arbitrary numbers of occlusion layers. We demonstrate in three publicly-available amodal datasets that the proposed method outperforms other layer-perception amodal segmentation and diffusion segmentation methods while producing reasonably diverse results. Future work will aim to augment efficiency and maintain output quality through super-resolution techniques and learned compression methods like VAEs. These advances will optimize our downsampling strategy, enabling a more efficient application to high-resolution datasets.

\bibliography{egbib}
\end{document}